\documentclass[10pt,twocolumn,letterpaper]{article}

\usepackage{iccv}
\usepackage{times}
\usepackage{epsfig}
\usepackage{graphicx}
\usepackage{amsmath}
\usepackage{amssymb}
\usepackage{array}

\usepackage[pagebackref=true,breaklinks=true,letterpaper=true,colorlinks,bookmarks=false]{hyperref}

\iccvfinalcopy 

\def\iccvPaperID{205} 

\DeclareMathOperator*{\argmin}{arg\,min}
\DeclareMathOperator*{\argmax}{arg\,max}

\ificcvfinal\fi
\begin{document}

\title{Colour Terms: a Categorisation Model Inspired by Visual Cortex Neurons}

\author{Arash Akbarinia\\
Centre de Visi\'{o} per Computador\\
Universitat Aut\`{o}noma de Barcelona\\
{\tt\small arash.akbarinia@cvc.uab.es}
\and
C. Alejandro Parraga\\
Centre de Visi\'{o} per Computador\\
Universitat Aut\`{o}noma de Barcelona\\
{\tt\small alejandro.parraga@cvc.uab.es}
}

\maketitle

\begin{abstract}
Although it seems counter-intuitive, categorical colours do not exist as external physical entities but are very much the product of our brains. Our cortical machinery segments the world and associate objects to specific colour terms, which is not only convenient for communication but also increases the efficiency of visual processing by reducing the dimensionality of input scenes. Although the neural substrate for this phenomenon is unknown, a recent study of cortical colour processing has discovered a set of neurons that are isoresponsive to stimuli in the shape of 3D-ellipsoidal surfaces in colour-opponent space. We hypothesise that these neurons might help explain the underlying mechanisms of colour naming in the visual cortex.

Following this, we propose a biologically-inspired colour naming model where each colour term -- e.g. red, green, blue, yellow, etc. -- is represented through an ellipsoid in 3D colour-opponent space. This paradigm is also supported by previous psychophysical colour categorisation experiments whose results resemble such shapes. ``Belongingness'' of each pixel to different colour categories is computed by a non-linear sigmoidal logistic function. The final colour term for a given pixel is calculated by a maximum pooling mechanism. The simplicity of our model allows its parameters to be learnt from a handful of segmented images. It also offers a straightforward extension to include further colour terms. Additionally, ellipsoids of proposed model can adapt to image contents offering a dynamical solution in order to address phenomenon of colour constancy. Our results on the Munsell chart and two datasets of real-world images show an overall improvement comparing to state-of-the-art algorithms.
\end{abstract}

\section{Introduction}

Colour vision contributes significantly to our perception of the world by providing valuable information about properties of objects and facilitating their segmentation from each other and the background~\cite{brainard2004color}. Its evolution might be guided by ecologically important tasks such as collecting ripe fruits or spotting predators. Besides that, our brains have evolved to communicate perception of colour through natural language. Colour terms are extensively practised in our day-to-day life. For instance, we tend to describe objects by their colour names (e.g. pass me the blue book; look at that orange house). Moreover, we explicitly benefit from colours to facilitate various tasks (e.g. software programmers colour-code their source code to aid interpretation; pedestrians and drivers rely on colour-coded city traffic lights). 

Consequently, any computer application seeking to intuitively interact with humans (e.g. visual searching, image labelling, and content retrieval) requires to embody colour naming in its routine~\cite{van2015overview}. Furthermore, numerous computer vision algorithms (such as scene segmentation, high-dynamic-range imaging (HDR), target tracking, object recognition, and texture classification) can greatly benefit from segmentation of an image to its constituent colours: either by improving their accuracy or lowering their computational complexity~\cite{van2015overview}. Despite the omnipresence of colour in our lives and the prominent role played by our perceptual machinery, only a handful of computational colour naming models has been developed and even fewer attempt to incorporate our knowledge of the perceptual system into them.

Colour naming (also referred here as ``colour categorisation'') is a highly multidisciplinary topic. A large-scale linguistic survey by anthropologists Berlin \& Kay~\cite{berlin1991basic} hinted at eleven basic colour terms -- i.e. black, blue, brown, green, grey, orange, pink, purple, red, white, and yellow -- that are shared across most evolved languages and cultures. Universality of these colour terms has been challenged by the role of linguistic contexts~\cite{hardin1997color}. Nevertheless, they have been reconfirmed in various other studies~\cite{boynton1990salience,sturges1995locating,kay2003resolving} and to a certain extent explained by physiological evidence that demonstrate low-level mechanisms contribute to colour categorical perception prior to language acquisition~\cite{siok2009language}. Present general consensus favours an intermediate free-from-language low-level colour perception stage supported by non-verbal cognitive experiments~\cite{indow1988multidimensional}.

Colour naming at first might appear to be fully deterministic (indeed a few computational models have taken this approach~\cite{tominaga1985colour,lin2001cross}). However, Kay \& McDaniel~\cite{kay1978linguistic} suggested that the determining perceptual input comes from the language-processing part of the brain. Therefore the underlying visual mechanism behind colour naming must be modelled by continuous mathematics, i.e. fuzzy logic. This insight (also supported by psychophysics) implies that in practice every pixel has a value of ``belongingness'' (from zero to hundred per cent) to each colour category which is directly computed from the measured reflectance spectrum of a surface at that point. 

Initial works on fuzzy models started with Lammens~\cite{ele1994computational}, who fitted the data collected by Berlin \& Kay into some variations of Gaussian functions. Mojsilovic~\cite{mojsilovic2005computational} continued this approach with a new perceptual colour metric. Seaborn et al.~\cite{seaborn2005fuzzy} clustered psychophysical colour points with a k-means algorithm while Benavente et al.~\cite{benavente2008parametric} tackled the problem by means of a triple-sigmoidal parametric model, with a few lightness planes sliced into different colour categories and the rest approximated through interpolation. Contrary to previous algorithms that are based on fitting colour categories to psychophysically obtained \textit{focal colours}, van de Weijer et al.~\cite{van2009learning} proposed to learn colour names from real-world images using probabilistic latent semantic indexing. Our proposal to capture colour terms using geometrical shapes is fundamentally different from current methods: (i) we benefit from parametric modelling~\cite{ele1994computational,benavente2008parametric} with the added advantage of partitioning the colour space directly into three-dimensional shapes rather than interpolating from two-dimensional planes; (ii) unlike algorithms that learn every pixel independently through histograms with no explicit constraints on colour regions~\cite{van2009learning}, we impose  ellipsoidal shapes that act as natural restrictions to such colour regions.

Acknowledging the fact that concept of colour is a product of our brain, it naturally follows that the best way to address colour naming is to model what we know physiologically and psychophysically about the human cortical machinery. For example, it is widely accepted that colour categorisation has been shaped by evolution and neonatal adaptation to break down an extremely complex world into cognitively tractable entities, reducing the nearly two million colours that can be distinguished perceptually~\cite{pointer1998number} to about thirty categories than can be recalled by average subjects~\cite{derefeldt1995colour}. In particular, the elven universal colour categories~\cite{berlin1991basic} are unlikely to be arbitrary and possibly reflect ideal divisions of an irregularly shaped perceptual color space~\cite{regier2007color}. In a recent psychophysical experiment Parraga \& Akbarinia~\cite{akbarinia2015biologically} observed that in chromatically opponent space categorical frontiers between these eleven universal colours form ellipsoidal shapes in line with the elliptical isorresponses of V1 neurons reported in a physiological study by Horwitz \& Hass~\cite{horwitz2012nonlinear}.

Following this rationale, in this paper we present a biologically-inspired colour naming model based on an ``ideal'' partitioning of colour-opponent space (as suggested by Regier et al.~\cite{regier2007color}) through parsimonious ellipsoidal shapes (as revealed by psychophysics~\cite{akbarinia2015biologically} and physiology~\cite{horwitz2012nonlinear}). We extend the work of~\cite{akbarinia2015biologically} by: (i) demonstrating that parameters of ellipsoids and growth ratio can be learnt more ecologically from segmented images; (ii) accounting for rotation along each axis and all ellipsoids; (iii) showing that it is straightforward to incorporate new colour terms within the new framework; (iv) prototyping means of ellipsoids adaptation to the image contents in order to account for the phenomenon of colour constancy; and (v) experimenting our model on real-world images.

\section{Ellipsoidal colour categorisation model}

In this section: (i) we review relevant physiological and psychophysical facts about colour vision and colour naming; (ii) we detail theory of proposed model; and (iii) we explain different means of obtaining parameters of our model.

\subsection{Colour perception}

\begin{figure}[ht]
\centering
	\includegraphics[width=\linewidth]{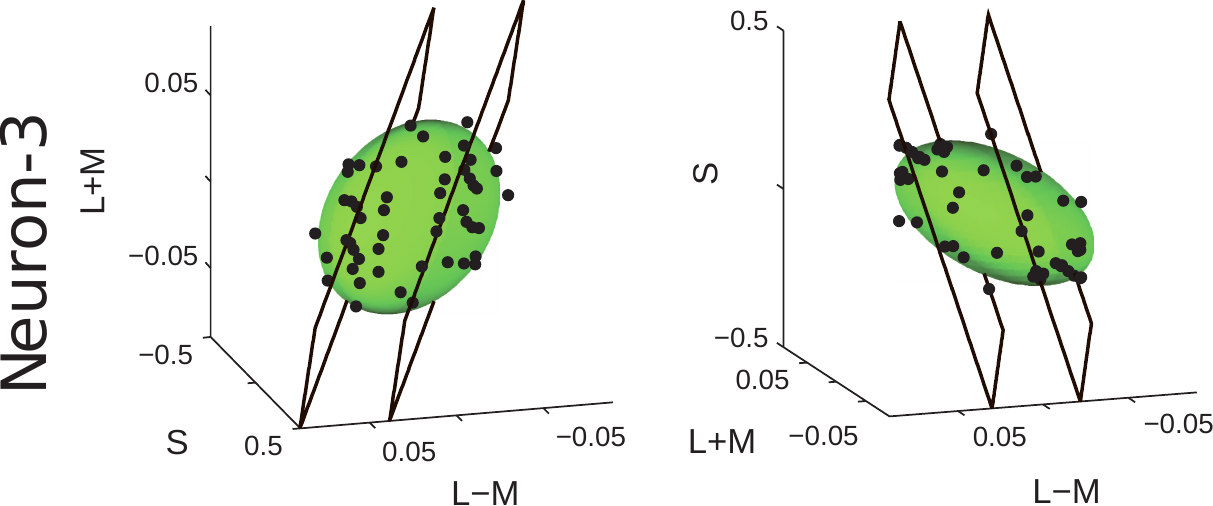}
	\caption{Two projections of Neurons-3 fitted to quadratic surfaces (green ellipsoids) in colour-opponent space, adapted from~\cite{horwitz2012nonlinear}. Black lines represent the best fitting planes.}
	\label{fig:handh}
\end{figure}

At present, we have a fairly rigorous understanding of cone photoreceptors that initiate colour vision by absorbing light at the back of retina. Signals produced by these cells are combined in an antagonistic manner to from the opponent channels that convey information to the visual cortex through the lateral geniculate nucleus (LGN)~\cite{fairchild2013color,parraga2013color}. As we advance deeper inside cortical areas, our knowledge of cerebral mechanisms involved in colour vision becomes less clear. In the primary visual cortex (V1), there is population of specialised neurons called single- and double-opponent cells that respond non-linearly to chromatic stimulus~\cite{shapley2011color}. A recent study by Horwitz \& Hass~\cite{horwitz2012nonlinear} analysed neurons in V1 in terms of their uniform responses to three-dimensional shapes in colour-opponent space. A large subset of these neurons (termed Neuron-3) responded best to ellipsoids whose major and minor axes are aligned to perceptual cardinal directions (see the schematics in Fig~\ref{fig:handh}). Their findings show how neurons of V1 can act jointly to process colour.

Similar ellipsoidal shapes have also emerged in psychophysical measures of colour boundaries where subjects were asked to produce the intermediate colour between two basic colour terms on a calibrated CRT monitor~\cite{akbarinia2015biologically}. This does not appear as a great surprise since colour categories tend to occupy connected regions of colour space~\cite{regier2007color}. However, these results could in turn  explain the organisation of universal colour terms around foci with perceptual constraints governing their position and shape, i.e. supporting the hypothesis that colour naming reflects optimal partitions of colour space~\cite{regier2007color}.

\subsection{Ellipsoidal partitioning of colour space (EPCS)}

We modelled each colour category as an ellipsoid in three-dimensional colour-opponent space following these rationale:
\begin{enumerate}
  \item The presence of Neuron-3 in V1~\cite{horwitz2012nonlinear} shows the plausibility of low-level models of opponent colour processing, i.e. ellipsoids are parsimonious shapes that can be implemented by low-level visual neurons.
  \item Contours of ellipsoids provide an appropriate fit to the psychophysically-measured colour categorical boundaries in~\cite{akbarinia2015biologically}.
  \item In this context, the centre of an ellipsoid can be interpreted as the focal colour and its geometrical properties determine the optimal partitioning~\cite{regier2007color}.
\end{enumerate}

An ellipsoid aligned to the axes of a Cartesian coordinate system is defined as:
\begin{equation}
  \left( \frac{x-x_0}{a} \right)^2  + \left( \frac{y-y_0}{b} \right)^2  +\left( \frac{z-z_0}{c} \right)^2 = 1 ,
\end{equation}
where $(x_0, y_0, z_0)$ are the coordinates of the ellipsoid-centre; and $(a, b, c)$ represent the length of the semi-axes. To account for any rotations around the axes of coordinate system, we defined our complete set of ellipsoid parameters $s$ with nine parameters:
\begin{equation}
  s = [(x_0, y_0, z_0), (a, b, c), (\theta, \phi, \gamma)] ,
\end{equation}
where $(\theta, \phi, \gamma)$ are the rotational angles around each of the colour-opponent axes.

A na\"{i}ve procedure to categorise pixels into different colour terms can be described as a simple binary test: when a pixel is inside an ellipsoid, it belongs to that category, otherwise it does not. However, there are two major flaws with this approach: (i) pixels outside of all ellipsoids will be categorised as neither of the colour terms; and (ii) the colour categorisation will lack the fuzziness proposed by~\cite{kay1978linguistic} as its underlying visual mechanism. Thus, to simulate the large variability present in the categorisation decision we utilised the sigmoid curve that is a spacial case of the logistic function, given as
\begin{equation}
 S(g) =  \frac{1}{1 + e ^ {-g}} ,
\end{equation}
where $g$ is the steepness of the curve (also knows as the growth ratio). Larger values of $g$ results in a more binarised categorisation, whereas smaller values of $g$ increase the fuzziness of our model.

There are various ways to model the steepness of each colour category. The simplest is to set $g$ as a constant number. Another strategy is to establish a relation between the steepness of each category and size of its ellipsoid. We favoured a more adjustable solution in which $g$ is set as a free variable for each colour category. This allows our model to vary its level of fuzziness for different colour names. Therefore, in our model each colour term, $t$, consists of ten parameters:
\begin{equation}
	t = [s; g] .
\end{equation}

Belongingness of a pixel to a colour category is computed by:
\begin{equation}
  B_t(x) = \frac{1}{1+e^{g_t(\vert p - c_t \vert - h)}},
  \label{eq:probability}
\end{equation}
where $B$ is the probability of pixel $p$ belonging to colour term $t$; $g_t$ is the steepness of the corresponding colour category; $c_t$ is the centre of its ellipsoid; $h$ is the position of the half-height transition point, which in our model is defined as the distance from the centre of an ellipsoid to its surface in the direction joining $c_t$ and $p$.

Although trivial, it is worth mentioning that when a pixel falls inside an ellipsoid, $\vert p - c_t \vert$ is smaller than $h$, as a result the input of the natural exponential function becomes a negative value. Consequently, the entire natural exponential term becomes smaller than $1$. The belongingness of a pixel to a colour category increases as $\vert p - c_t \vert - h$ tends towards $-\infty$ and it reaches its maximum value at the centre of an ellipsoid, where the exponential term drops to $0$.

Deterministic colour naming requires a unique term for every pixel. This can be achieved through different strategies of combining probabilities of all colour categories, for instance considering the perceptually neighbouring colour (i.e. red and orange, or pink and purple). However, this is beyond the scope of this paper and we adopted a simple maximum pooling mechanism: the highest probability among all colour categories is assigned as the colour term $C$ of that pixel:
\begin{equation}
  C(x) = \argmax_{t}B_t(x)
\end{equation}

\subsection{Acquiring model parameters}

\subsubsection{Colour space} The first prerequisite for modelling the processes that occur in the visual cortex is to represent the chromatic signal in a colour-opponent space (resembling the signal arriving from the retina). We selected the CIE L*a*b* colour space because is considered to be perceptually uniform~\cite{cie2004} and is widely used in computer vision and visual sciences. Nevertheless, since in our model we employ ellipsoids to partition a given colour space into different colour categories, our model is not dependant on the CIE L*a*b* and should work equally well in other colour-opponent spaces, such as CIE L*u*v*, lsY and DKL.

\subsubsection{Parameters optimisation}
\label{sec:paropm}
The proposed colour ellipsoids are parsimonious geometrical shapes whose parameters can be determined by different procedures. The simplest option would be to draw those ellipsoids manually and set the steepness to a constant value. Alternatively, surface of each ellipsoid can be fitted into data points that represent boundaries of a colour term; and the steepness of a category can be defined as the average length of its ellipsoid semi-axes, $g_t=\frac{a_t+b_t+c_t}{3}$, similar to~\cite{akbarinia2015biologically}. The most comprehensive solution would probably be to construct a ground truth for every point in a canonical colour-opponent space by means of psychophysical experiments. From this ground truth all the ten parameters of our model can simultaneously be learnt in an optimisation framework.

However, in practice collecting such an exhaustive ground truth from a large set of subjects is extremely time consuming. To overcome this issue we simulated the ground truth from the validation set of the Ebay colour naming dataset presented in~\cite{van2009learning} (8 images per each of the eleven basic colour names, making a total of 88). Given pixel $p$, we counted the number of times it was categorised as each of the eleven basic colour names. Dividing this by the total number of times pixel $p$ was categorised resulted in the degree of membership to each colour term.

We learnt the parameters of our model with a sequential quadratic programming optimisation method ($10^3$ number of iterations and $10^{-3}$ as tolerance constraint) with the error function
\begin{equation}
  \argmin \sum_{x=1}^{N} B_t(x) - G_t(x),
\end{equation}
where $N$ is number of pixels in the ground truth set; $B_t$ is defined in Eq.~\ref{eq:probability}; and $G_t(x)$ is the ground truth value of pixel $x$ belonging to category $t$. We simply initialised each colour ellipsoid, $i_t$, as follows
\begin{equation}
  i_t = [(\mu_{L*t}, \mu_{a*t}, \mu_{b*t}), (10, 10, 10), (0, 0, 0); 1]
\end{equation}
where $(\mu_{L*t}, \mu_{a*t}, \mu_{b*t})$ are the average coordinates (in CIE L*a*b* colour space) of all the pixels whose ground truth value of category $t$ is non-zero. We did not set any constraints on the optimisation of ellipsoid centres. Naturally, we restricted the length of semi-axes to positive values and the rotational angles to the range of $[0, \pi)$. Steepness of sigmoidal function was limited to the range of $(0, 1]$. 

\begin{figure}[ht]
\centering
 \includegraphics[width=0.98\linewidth]{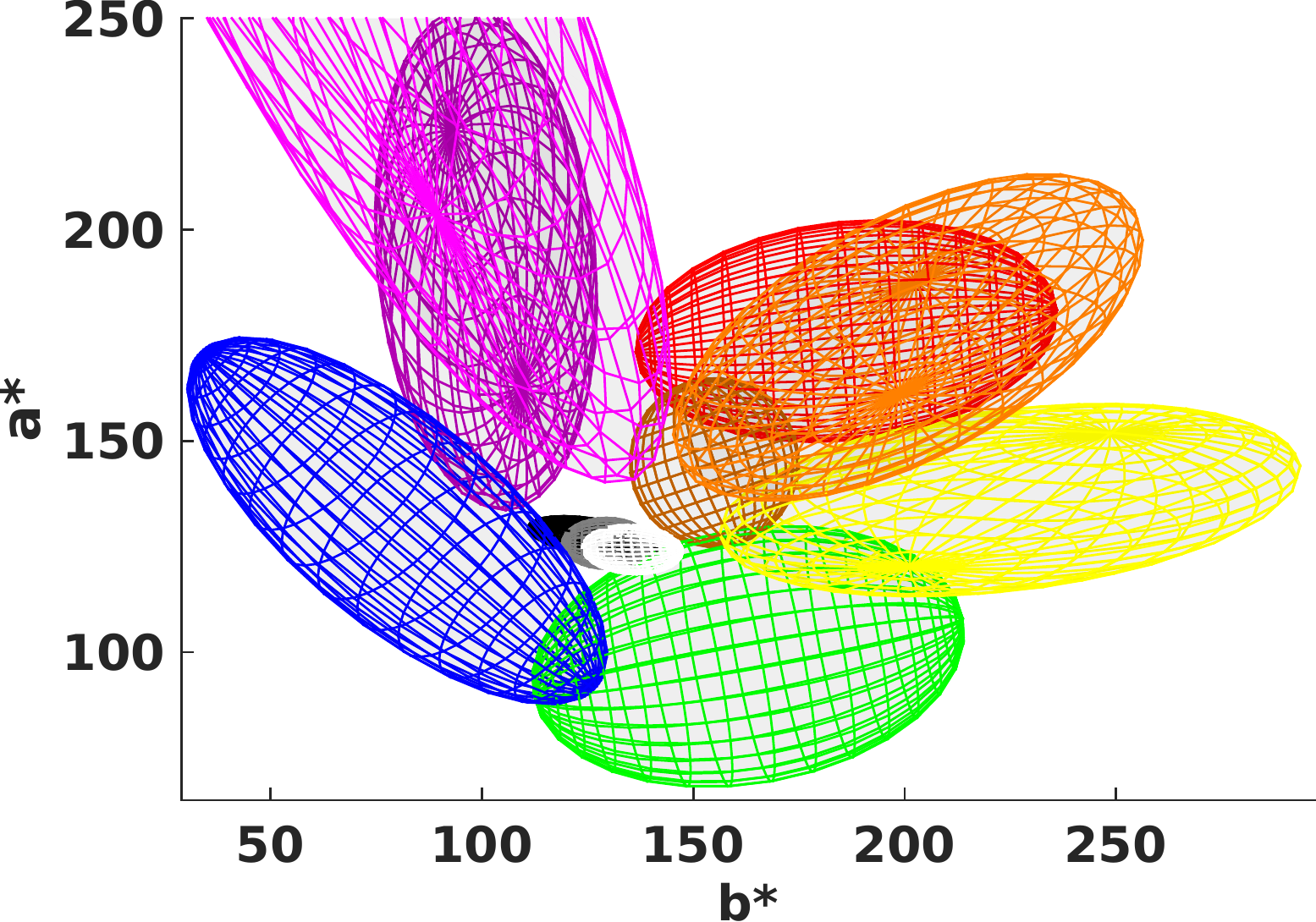}
\caption{Zenithal view (the $L=0$ plane) of the learnt colour ellipsoids (each ellipsoids corresponds to one of the eleven basic colour terms) in the CIE L*a*b* colour space.}
\label{fig:ellipsoids}
\end{figure}

Fig~\ref{fig:ellipsoids} illustrates the eleven colour ellipsoids learnt from our simulated ground truth. One can highlight a few aspects of the zenithal view that express high congruence with our very own colour perception as follow:
\begin{itemize}
  \item The achromatic categories are placed at the centre of all ellipsoids in line with the hue circle, first proposed by Newton.
  \item The ellipsoids corresponding to opponent colours, i.e. red-green and yellow-blue, do not overlap. This is in line with Hering's colour theory which states that these colour cannot be perceived together.
\end{itemize}

\section{Experiments and results}

We learnt parameters of our model -- termed Ellipsoidal Partitioning of Colour Space (EPCS) -- from two different ground truths:
\begin{itemize}
  \item \textit{EPCS [Rw]} -- learnt only from real-world images by extracting the ground truth from validation set of~\cite{van2009learning}.
  \item \textit{EPCS [Ps]} -- to account for colour naming experiments we averaged pixel probabilities of real-world ground truth with the psychophysical results of~\cite{akbarinia2015biologically}.
\end{itemize}

We quantitatively evaluated the proposed model by conducting experiments on two different kinds of datasets: (i) colour chips categorised by psychophysical experiments; and (ii) colour segmented objects in real-world images.

\begin{figure*}
\centering
\begin{tabular}{ccc}
 Munsell chart & EPCS [Ps] Segmentation & EPCS [Rw] Segmentation \\
  \includegraphics[width=0.31\linewidth]{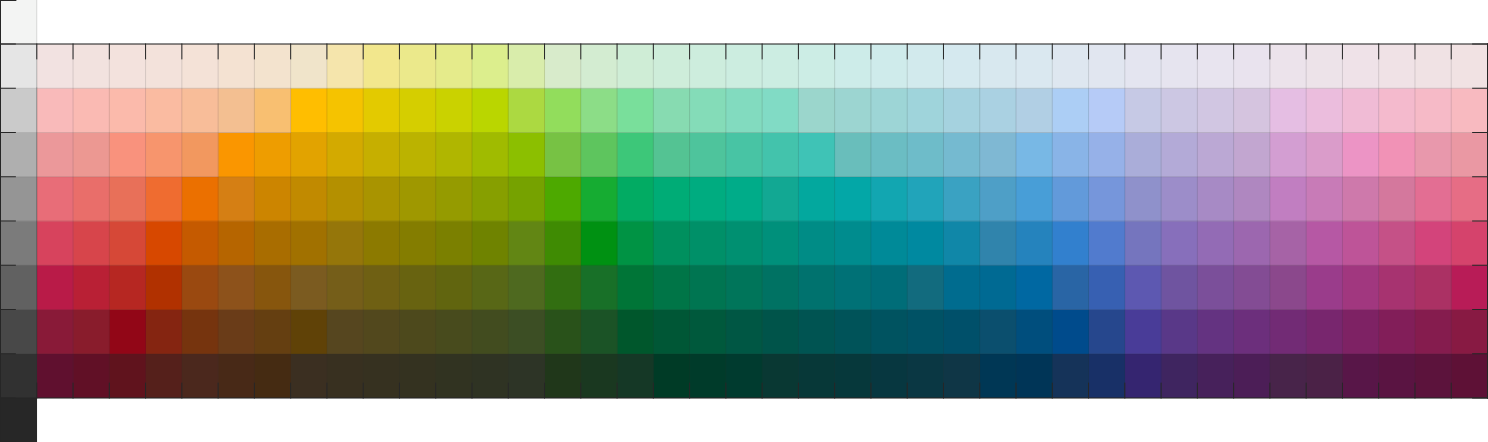} & \includegraphics[width=0.31\linewidth]{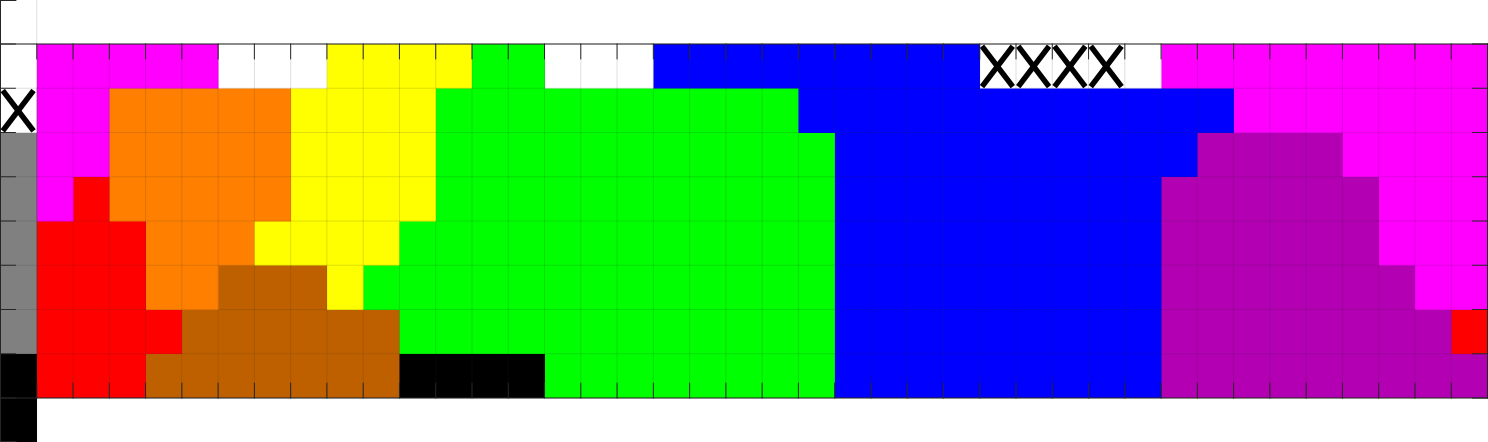} & \includegraphics[width=0.31\linewidth]{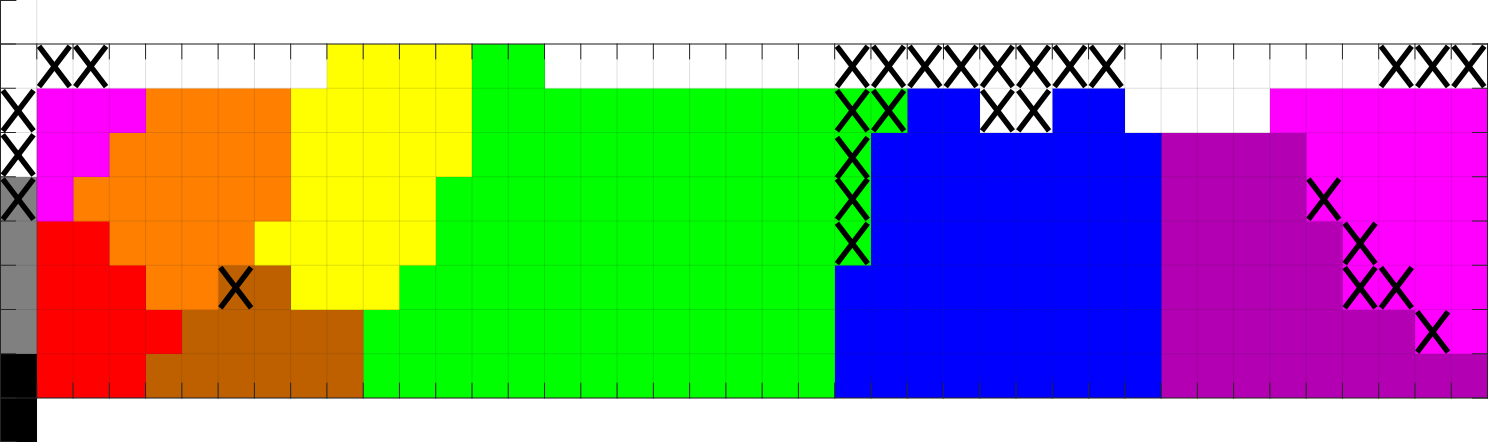} 
\end{tabular}
\caption{Result of EPCS applied to the Munsell colour chart.}
\label{fig:munsellresults}
\end{figure*}

\subsection{Munsell colour chart}

The left panel of Fig~\ref{fig:munsellresults} shows the Munsell chart that contain 330 different pixels (eight chromatics rows, each consisting of 40 hues in increments of 2.5, and one column of 10 achromatic lightness). Many colour naming studies have compared their categorisation results to the psychophysical experiments of Berlin \& Kay~\cite{berlin1991basic} (24 native speakers from 110 languages were asked to name each Munsell chip) and Sturges \& Whitfield~\cite{sturges1995locating} (20 English speakers named each Munsell sample twice). Our segmentation of the Munsell chart is illustrated on the right panel of Fig~\ref{fig:munsellresults}. Our results match perfectly with psychophysical experiment of Sturges \& Whitfield and only vary on five points (all caused by the white colour) comparing to the survey of Berlin \& Kay.

Table~\ref{tab:munsresults} quantitatively compares accuracy of our model to seven state-of-the-art algorithms that have also reported theirs results on the Munsell chart. In comparison to the colour naming survey of Berlin \& Kay~\cite{berlin1991basic} EPCS practically matches the best reported results in the literature (NICE), far ahead of the third best colour naming models (SFKM, TSEM). With respect to the psychophysical experiment of Sturges \& Whitfield~\cite{sturges1995locating} our model along with SFKM, TSEM and NICE obtains perfect accuracy.

We can observe a large difference between two variations of our model mainly caused by white pixels. EPCS [Rw] (learnt only from real-world images) categorises pixels with a faint colour as white, whereas EPCS [Ps] (learnt by influence of colour naming experiments in controlled environment) categorise those pixels into chromatic categories. This is an issue noted by~\cite{van2009learning} as well.

\begin{table}[ht]
\centering
\begin{tabular}{|>{\raggedright\arraybackslash}m{1.0cm}@{\hskip 0.04in}>{\raggedright\arraybackslash}m{0.7cm}@{\hskip 0.04in}|>{\centering\arraybackslash}m{2.57cm}@{\hskip 0.04in}|>{\centering\arraybackslash}m{2.82cm}|}
\cline{3-4}
\multicolumn{2}{c|}{} & Berlin \& Kay & Sturges \& Whitfield \\ \cline{3-4}
\hline
LGM & \cite{ele1994computational} & 0.77 & 0.83  \\
MES & \cite{maclaury1992brightness} & 0.87 & 0.96 \\
TSM & \cite{benavente2004fuzzy} & 0.88 & 0.97 \\
SFKM & \cite{seaborn2005fuzzy} & 0.92 & 1.00 \\
TSEM & \cite{benavente2008parametric} & 0.92 & 1.00 \\
PLSA & \cite{van2009learning} & 0.89 & 0.98 \\
NICE & \cite{akbarinia2015biologically} & 0.98 & 1.00 \\
\hline
\textbf{EPCS} & [Ps] & \textbf{0.98} & \textbf{1.00} \\
\textbf{EPCS} & [Rw] & \textbf{0.87} & \textbf{0.98} \\
\hline
\end{tabular}
\caption{True positive ratio of several colour naming models on psychophysical experiments. Lammens's Gaussian model (LGM)~\cite{ele1994computational}, MacLaury's English speaker model (MES)~\cite{maclaury1992brightness}, Benavente \& Vanrell's triple sigmoid model (TSM)~\cite{benavente2004fuzzy}, Seaborn's fuzzy k-means model (SFKM)~\cite{seaborn2005fuzzy}, Benavente et al.'s triple sigmoid elliptic model (TSEM)~\cite{benavente2008parametric}, van de Weijer et al.'s probabilistic latent semantic analysis (PLSA)~\cite{van2009learning}, Parraga \& Akbarinia's neural isoresponsive colour ellipsoids (NICE), and the proposed ellipsoidal partitioning of colour space (EPCS).}
\label{tab:munsresults}
\end{table}

\subsection{Real-world images}

We evaluated the proposed model on two datasets of real-world images\footnote{The source code and all the materials are available under this link \url{https://goo.gl/ZCBLJA}.}. Along with our model we tested three state-of-the-art methods (whose source codes are publicly available) : Benavente et al.'s triple sigmoid elliptic model (TSEM)~\cite{benavente2008parametric}, van de Weijer et al.'s probabilistic latent semantic analysis (PLSA)~\cite{van2009learning}, and Parraga \& Akbarinia's neural isoresponsive colour ellipsoids (NICE)~\cite{akbarinia2015biologically}. We assessed each algorithm based on their true positive ratio, i.e. $\frac{TP}{TP + FN}$, where $TP$ represents pixels whose colour names are correctly labelled and $FN$ are those that are mislabelled. Due to the nature of the available ground truths, which primarily contain one colour category per image, other evaluation metrics were inappropriate. Images of tested datasets are of various size and in order to avoid the bias for smaller images, we first computed the true positive ratio for each image and reported results are averaged over all.

\subsubsection{Ebay dataset}

Ebay dataset~\cite{van2009learning} consists of four sets of man-made objects, i.e. cars, dresses, pottery and shoes. Every set contains 110 images, i.e. ten images for each of the eleven basic colour terms. The ground truth masks are based on semi-automatic segmentation algorithms. To compensate for absence of natural objects (such as fruits, vegetables, flowers etc. -- where colour arguably plays an important role for recognition) we extended this dataset by creating an extra set of images containing natural objects following the same procedure as the original authors.

We have reported true positive ratio of four methods on Ebay dataset in Table~\ref{tab:ebayresults}. Evidently EPCS [Rw] outperforms all other methods. We can also observe a large gap between performance of EPCS [Ps] and PLSA in comparison to TSEM and NICE in all five subcategories. In three sets (dresses, shoes and natural) EPCS [Ps] obtains higher true positive ratio compared to PLSA. Advantage of EPCS [Ps] over PLSA becomes more tangible by considering their respective performance on psychophysical data, where EPCS [Ps] performs notably better (see Table~\ref{tab:munsresults}).

\begin{table}[ht]
\centering
\begin{tabular}{|l@{\hskip 0.04in}l@{\hskip 0.04in}|c@{\hskip 0.04in}|c@{\hskip 0.04in}|c@{\hskip 0.04in}|c@{\hskip 0.04in}|c@{\hskip 0.04in}|}
\cline{3-7}
\multicolumn{2}{c|}{} & Cars & Dresses & Pottery & Shoes & Natural \\
\hline
TSEM & \cite{benavente2008parametric} & 0.59 & 0.68 & 0.62 & 0.73 & 0.69 \\
PLSA & \cite{van2009learning} & 0.60 & 0.82 & 0.76 & 0.78 & 0.77  \\
NICE & \cite{akbarinia2015biologically} & 0.52 & 0.69 & 0.54 & 0.67 & 0.67 \\
\hline
\textbf{EPCS} & [Ps] & \textbf{0.60} & \textbf{0.84} & \textbf{0.76} & \textbf{0.79} & \textbf{0.80} \\
\textbf{EPCS} & [Rw] & \textbf{0.65} & \textbf{0.86} & \textbf{0.80} & \textbf{0.80} & \textbf{0.80} \\
\hline
\end{tabular}
\caption{True positive ratio of four colour naming models on Ebay dataset for each subcategory.}
\label{tab:ebayresults}
\end{table}

\subsubsection{Small objects dataset}

Small objects dataset \cite{yuan2015illumination} contains 300 16-bit images of various material (e.g. paper, plastic, metal, wood, fruits) captured under different types of illuminants. Each image comes with a manual segmentation of its constituting regions according to their colour names. However, number of pixels for each of the eleven basic colour terms is not uniformly distributed.

We have reported true positive ratio of four methods on small objects dataset in Table~\ref{tab:smallresults}. We can observe similar scenarios as Ebay dataset. EPCS [Rw] performs best among all while EPCS [Ps] and PLSA obtain better results in comparison to TSEM and NICE.

\begin{table}[ht]
\centering
\begin{tabular}{|ll|c|}
\cline{3-3}
\multicolumn{2}{c|}{} & Small Objects \\
\hline
TSEM & \cite{benavente2008parametric} & 0.69 \\
PLSA & \cite{van2009learning} & 0.73  \\
NICE & \cite{akbarinia2015biologically} & 0.52 \\
\hline
\textbf{EPCS} & [Ps] & \textbf{0.73} \\
\textbf{EPCS} & [Rw] & \textbf{0.77} \\
\hline
\end{tabular}
\caption{True positive ratio of four colour naming models on small objects dataset.}
\label{tab:smallresults}
\end{table}

\section{Discussion}

\begin{figure}[ht]
\centering
\begin{tabular}{c@{\hskip 0.04in}c@{\hskip 0.04in}c}
 Original & EPCS [Ps] & EPCS [Rw] \\
  \includegraphics[width=0.31\linewidth]{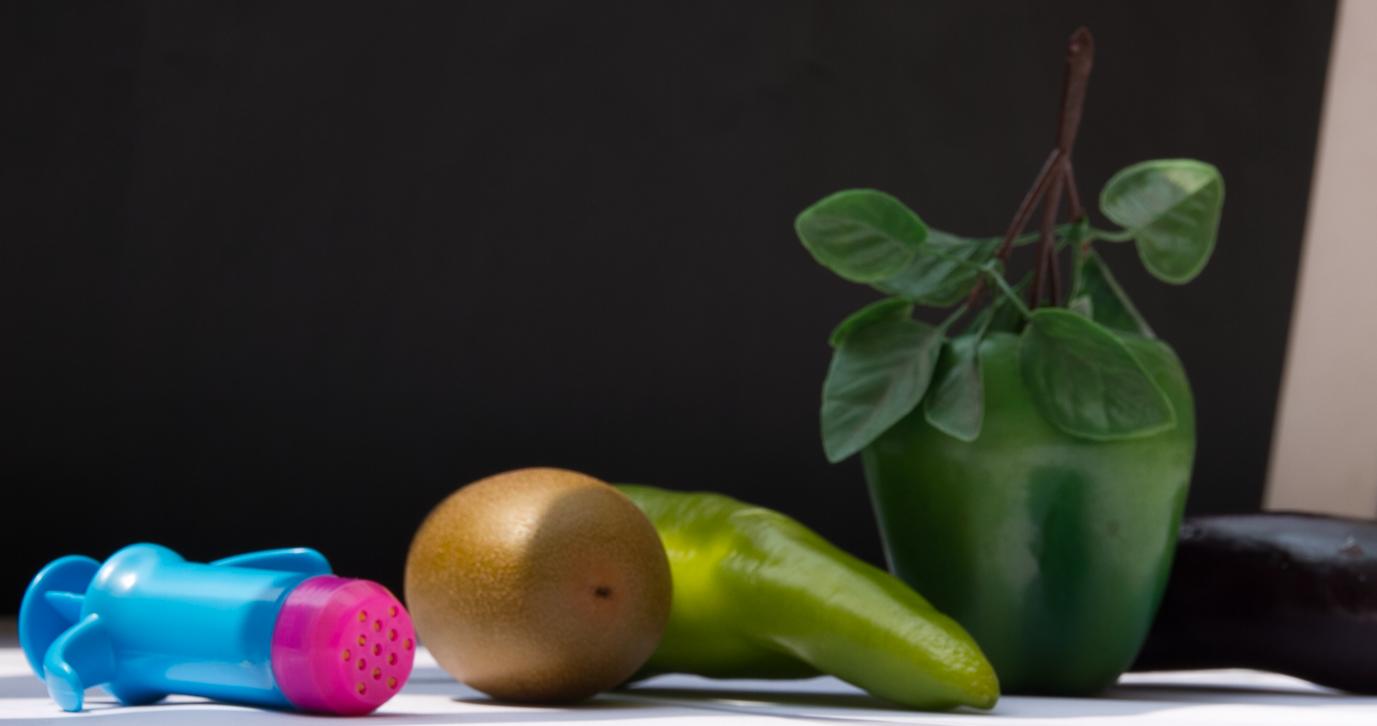}
  &
  \includegraphics[width=0.31\linewidth]{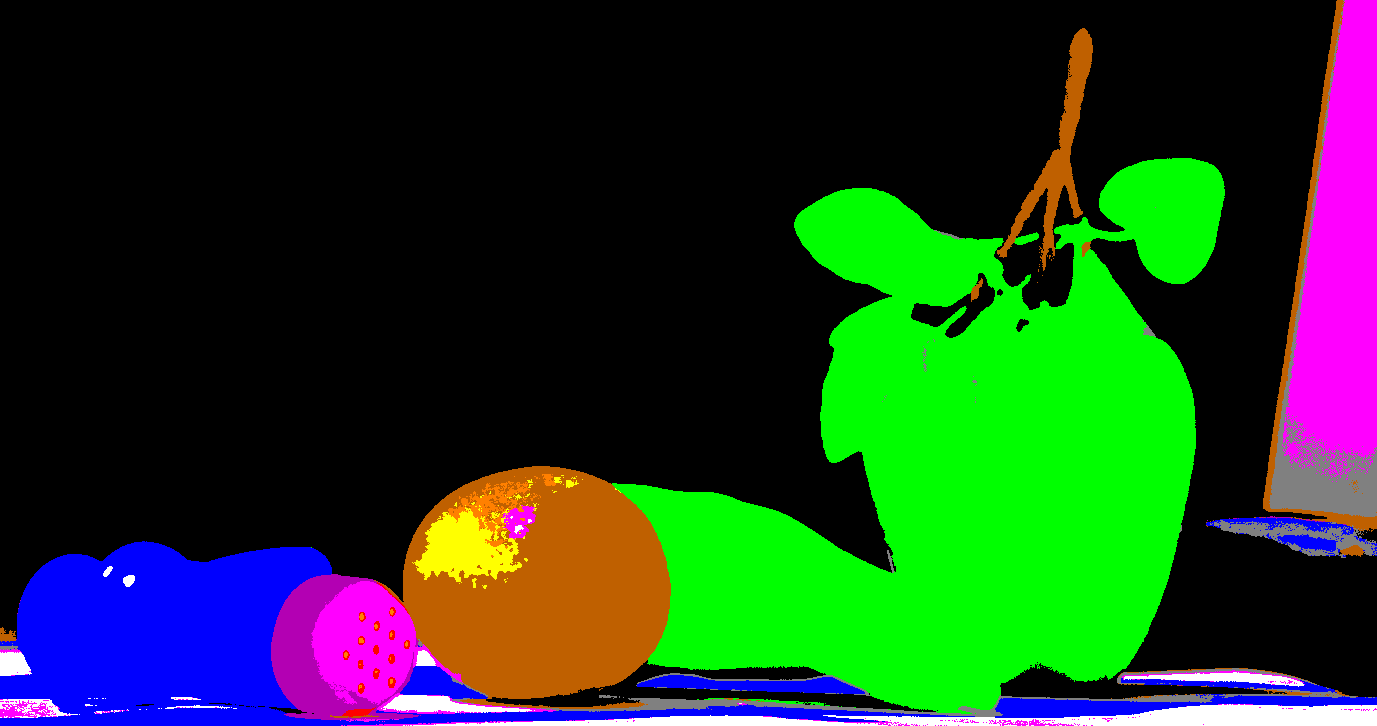}
  &
  \includegraphics[width=0.31\linewidth]{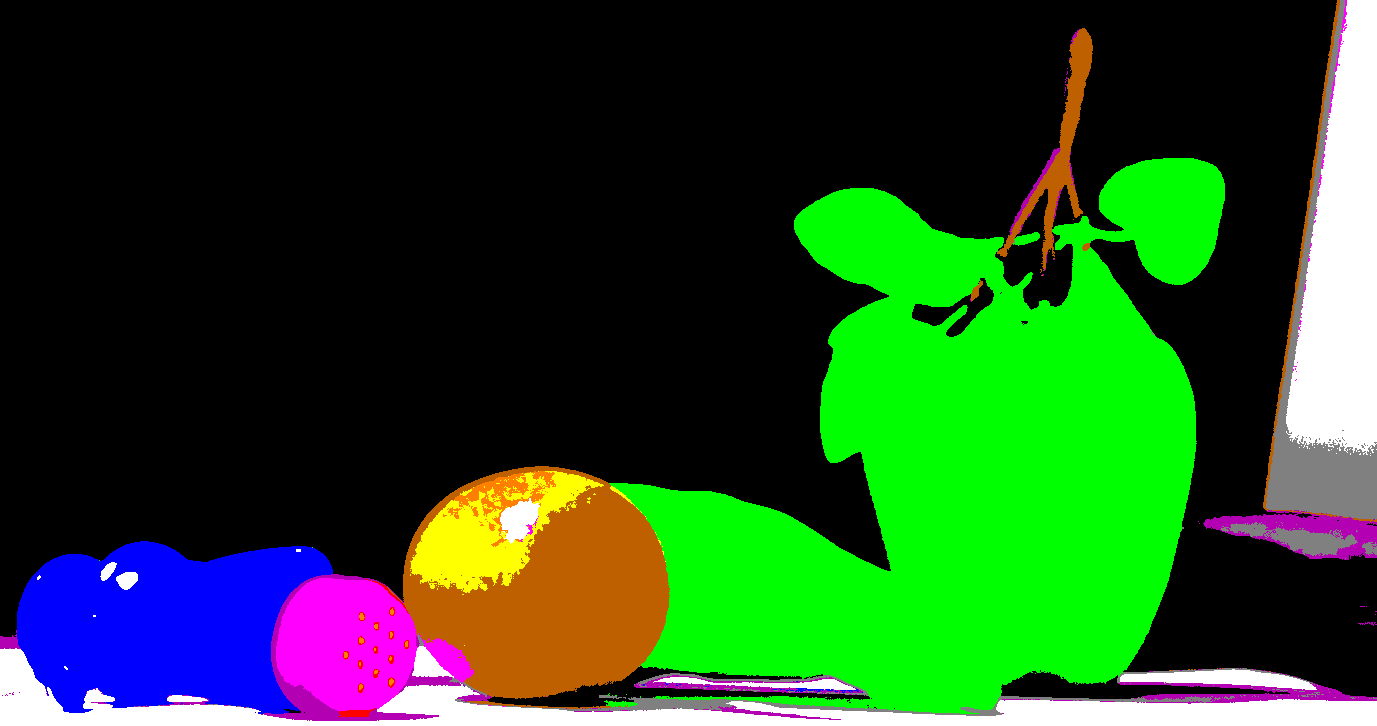}
\end{tabular}
\caption{EPCS [Ps] versus [Rw] in a real-world image.}
\label{fig:psrw}
\end{figure}

We have illustrated one exemplary image from the small object dataset in Fig~\ref{fig:psrw}. We can observe that EPCS [Ps] mislabels the white part of the wall as pink. This is the main reason that EPCS [Rw] clearly performs better than EPCS [Ps] in real-world images (refer to Tables \ref{tab:ebayresults} and \ref{tab:smallresults}). However, we would like to emphasise results of the later one that in psychophysical data where environment is controlled and no noise is present obtain almost perfect true positive ratio (similar to NICE), and it reasonably performs well on real-world images (contrary to NICE). We believe high accuracy on psychophysical experiments is essential because a colour naming model should first and foremost correctly categorise individual pixels. Other challenging tasks of colour naming, e.g. faint colours appear as white in an image context, are caused by different phenomenons such as colour constancy and induction. These challenges should be solved by modelling colour naming in a dynamic fashion.

\begin{figure*}
\centering
\begin{tabular}{c@{\hskip 0.04in}|@{\hskip 0.04in}c@{\hskip 0.04in}|@{\hskip 0.04in}c}
  \includegraphics[width=0.32\linewidth]{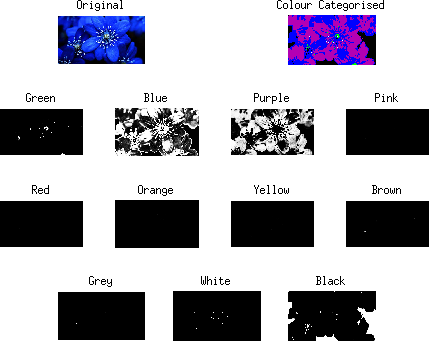} & \includegraphics[width=0.32\linewidth]{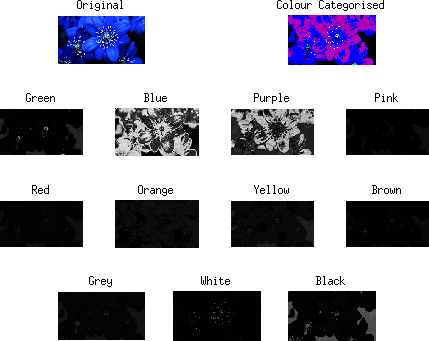} & \includegraphics[width=0.32\linewidth]{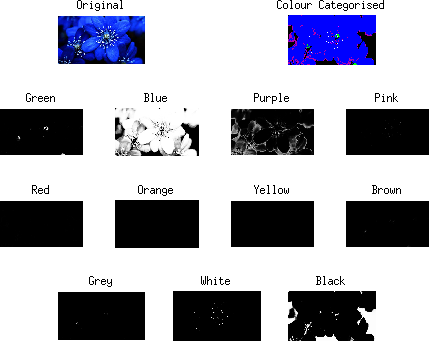} \\
  \hline
  & & \\
  \includegraphics[width=0.32\linewidth]{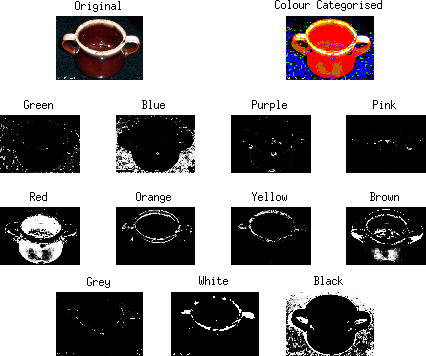} & \includegraphics[width=0.32\linewidth]{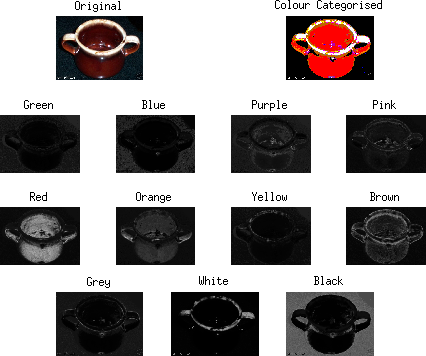} & \includegraphics[width=0.32\linewidth]{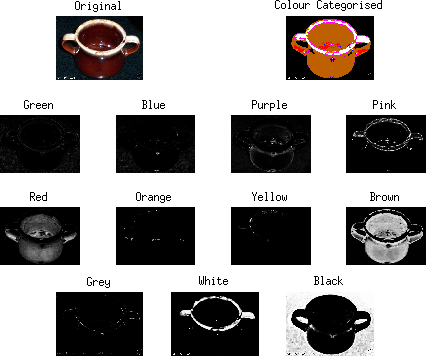} \\
  \hline
  & & \\
  \includegraphics[width=0.32\linewidth]{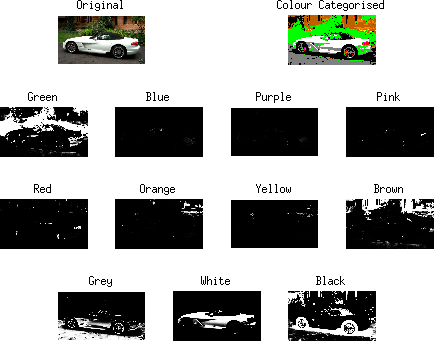} & \includegraphics[width=0.32\linewidth]{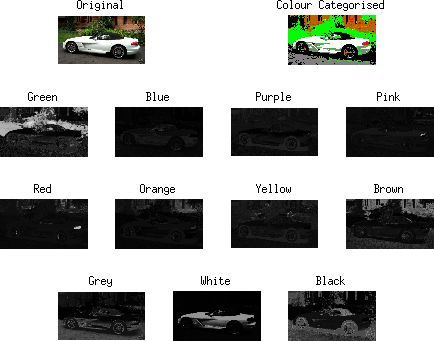} & \includegraphics[width=0.32\linewidth]{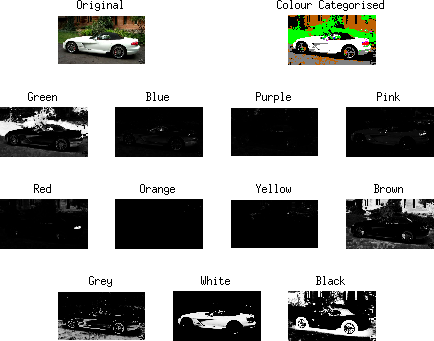} \\
  (a) TSEM~\cite{benavente2008parametric} & (b) PLSA~\cite{van2009learning} & (c) \textbf{EPCS} [Ps]
\end{tabular}
\caption{Detailed comparison of three different algorithms on real-world images. Individual panels consists of the original image, its colour categorised version, and elven probability maps corresponding to each basic colour term.}
\label{fig:comparsiondetails}
\end{figure*}

Fig~\ref{fig:comparsiondetails} shows three examples from the real-world datasets. In each panel the original image is displayed accompanied with its respective results from each of the algorithms considered: TSEM~\cite{benavente2008parametric}, PLSA~\cite{van2009learning}, and EPCS. The blue flowers (first row) which are largely misclassified as purple by TSEM and PLSA, are correctly assigned to the blue category in our model. We observed quite a few similar cases with other blue objects.

The brown pottery mug (Fig~\ref{fig:comparsiondetails}, second row) is almost entirely miscategorised as red by TSEM and PLSA. On the contrary, EPCS accurately labels it as brown. A closer inspection to the corresponding probability maps reveals that TSEM assigns pixels of the mug to the red category with a very high probability (almost $100\%$). PLSA labels them as red (with $60\%$ probability) while granting some likelihood to perceptually neighbouring colours (about $20\%$ to orange and $10\%$ to brown). However, this uncertainty spreads to the purple category as well with about $5\%$ probability. EPCS's results show more consistency with about $60\%$ probability on the brownness of the mug, while acknowledging that the neighbouring colour red is also probable (with about $40\%$). It is worth paying extra attention to the background as well, where TSEM misassigns a great portion of it to the blue category. Contrary to this, PLSA and EPCS have no difficulties to correctly label it as black.

The white car, shown in the third row of Fig~\ref{fig:comparsiondetails}, is a difficult case due to the cast of green light over its body and surroundings. All three methods, in general, accurately label the car as white. Nonetheless, there are pixels near the back wheel and on the front door that are mistakenly categorised as green. This issue is more noticeable for TSEM and its minimal in EPCS.

\subsection{Model extension} 
\begin{figure}[ht]
\centering
\begin{tabular}{cc}
  \includegraphics[width=0.34\linewidth]{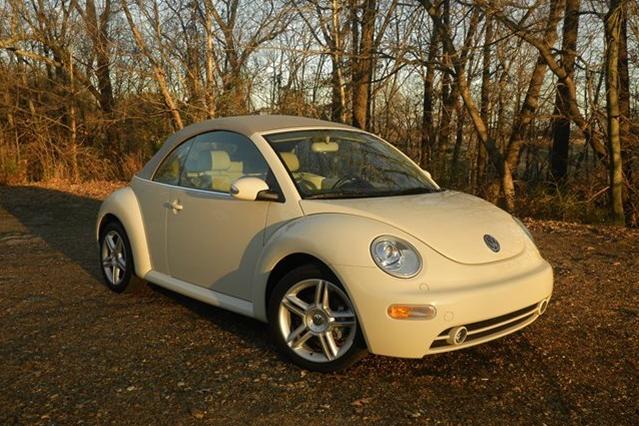}
  &
  \includegraphics[width=0.34\linewidth]{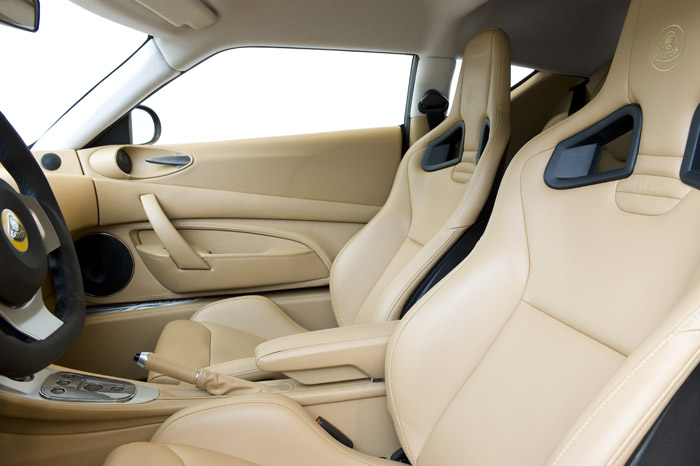}
\end{tabular}
\caption{Two images used in learning colour cream.}
\label{fig:learncream}
\end{figure}

There are situations where one might want to add extra categories to the eleven basic colour terms (e.g., some languages contain two names for ``blue'' like Russian, Italian and Spanish from the River Plate area). Alternatively, there are many intermediate colour terms used in everyday language (such as, olive, turquoise, cream) that arguably deserve their own category. New colour names are usually learnt by humans after the presentation of very few examples and our model can simulate this process straightforwardly. As an illustration, we learnt the colour term ``cream'' from merely two images, (see Fig~\ref{fig:learncream}), following the same procedure explained in section~\ref{sec:paropm}.

\begin{figure}[ht]
\centering
\begin{tabular}{c@{\hskip 0.04in}c@{\hskip 0.04in}c}
 Original  & 11 Categories & 12 Categories \\
\includegraphics[width=0.315\linewidth]{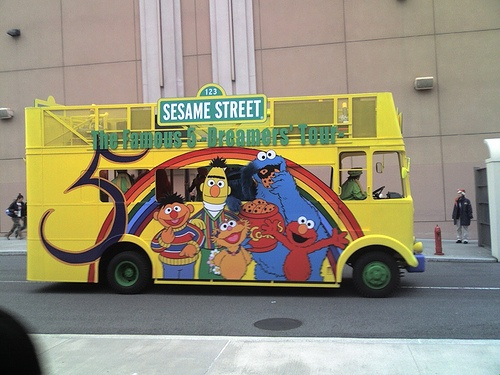} & \includegraphics[width=0.315\linewidth]{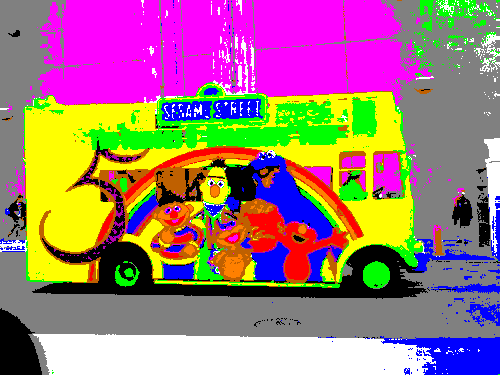} & \includegraphics[width=0.315\linewidth]{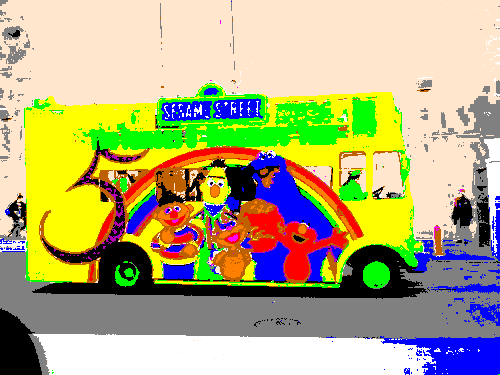}
\end{tabular}
\begin{tabular}{c@{\hskip 0.04in}c@{\hskip 0.04in}c@{\hskip 0.04in}c}
  Green & Blue & Purple & Pink \\
  \includegraphics[width=0.23\linewidth]{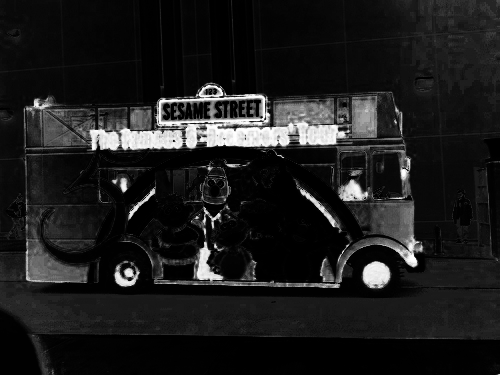} & \includegraphics[width=0.23\linewidth]{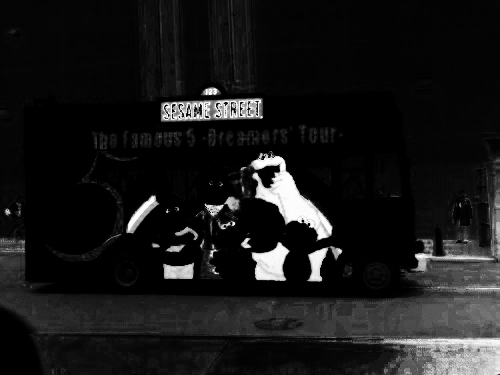} & \includegraphics[width=0.23\linewidth]{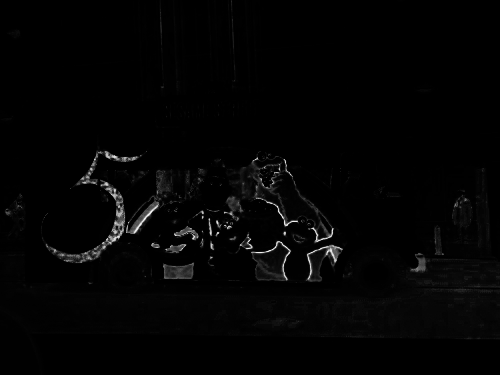} & \includegraphics[width=0.23\linewidth]{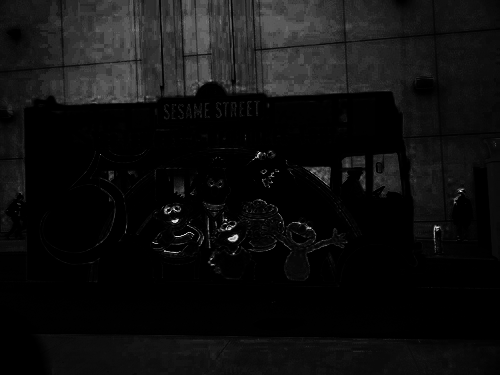} \\
  Red & Orange & Yellow & Brown \\
  \includegraphics[width=0.23\linewidth]{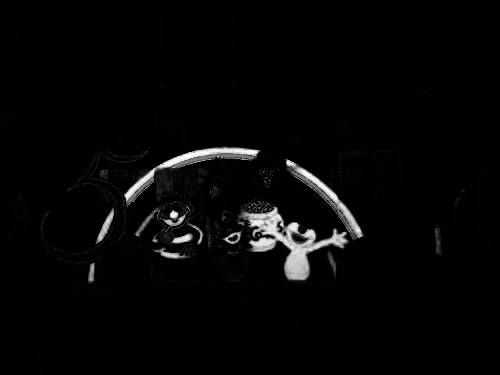} & \includegraphics[width=0.23\linewidth]{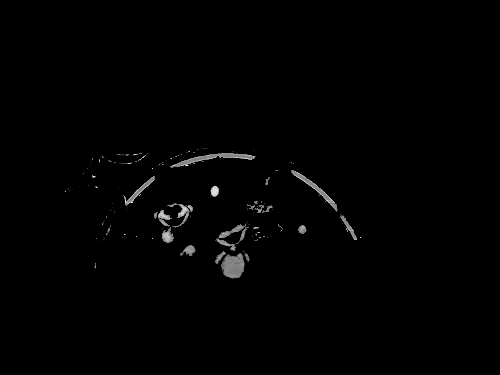} & \includegraphics[width=0.23\linewidth]{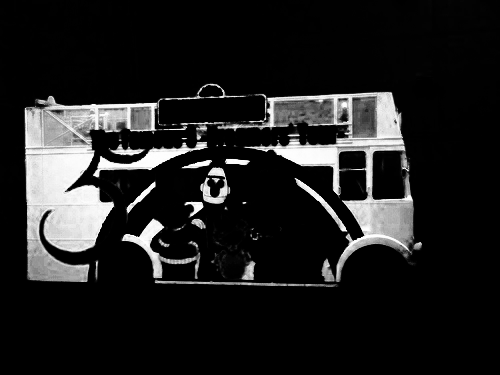} & \includegraphics[width=0.23\linewidth]{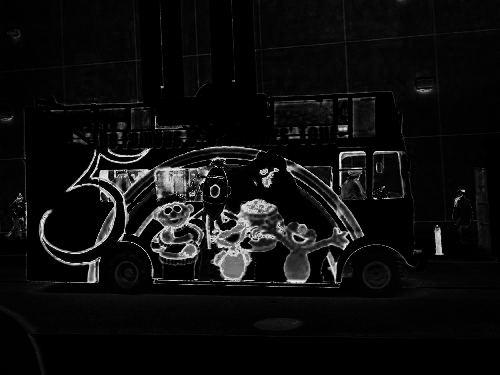} \\
  Grey & White & Black & Cream \\
  \includegraphics[width=0.23\linewidth]{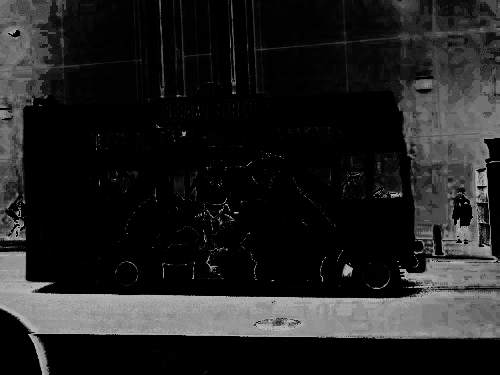} & \includegraphics[width=0.23\linewidth]{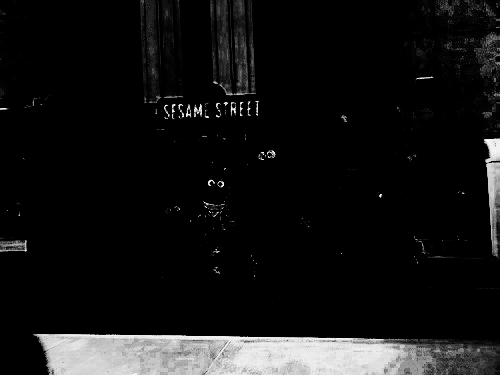} & \includegraphics[width=0.23\linewidth]{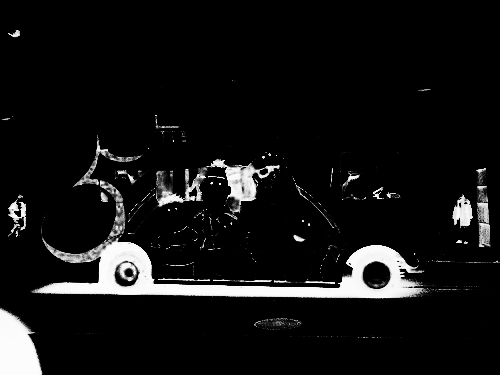} &
\includegraphics[width=0.23\linewidth]{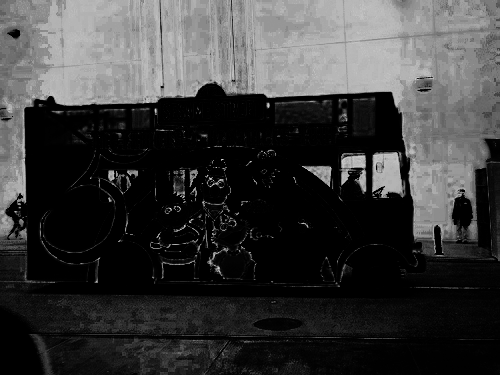}
\end{tabular}
\caption{Colour categorisation including an extra category for the cream colour. Top row shows the original image with is respective segmentation according to eleven or twelve colour categories.}
\label{fig:creamcolour}
\end{figure}

Fig~\ref{fig:creamcolour} shows the impact of this newly introduced cream category on the colour segmentation of an image from the Pascal Project Dataset~\cite{everingham2010pascal}. Relying only on the eleven colour terms EPCS incorrectly labels the wall on the back of the image as pink (although with low probability that is on average smaller than $20\%$). Segmenting with twelve categories allows us to accurately classify the wall as cream. The flexibility of our algorithm can be further exploited to create a personalised colour naming model which reflects the individual variability present in the psychophysical data. This is very economical and can be achieved by segmenting a handful of images from a personal digital assistant (PDA) for example. Furthermore, an interactive application can allow subjects to manipulate the colour ellipsoids directly to achieve the colour categorisation they desire.

\subsection{Model adaptation}
One important aspect of any colour naming model is its context adaptability. This is feasible within our model by dynamically adjusting the ellipsoids to the image or even the pixel being processed. One of the greatest challenges in colour naming algorithms is the frontier between chromatic and achromatic colours as we experienced in our experiments and mentioned by~\cite{van2009learning}. In a neutral background colours appear more saturated comparing to a colourful environment~\cite{brown1997color}. As a proof-of-concept we attempted to address this issue by adapting achromatic ellipsoids to the level of colourfulness of an image. We stretched the chromatic semi-axes (a*b*) of achromatic ellipsoids on the direction that average pixels of an image differed from neutral grey. The results of this experiment are reported in Table~\ref{tab:daptation}. 

\begin{table}[ht]
\centering
\begin{tabular}{|@{\hskip 0.05in}c@{\hskip 0.05in}|@{\hskip 0.05in}c@{\hskip 0.05in}|@{\hskip 0.05in}c@{\hskip 0.05in}|@{\hskip 0.05in}c@{\hskip 0.05in}|@{\hskip 0.05in}c@{\hskip 0.05in}|@{\hskip 0.05in}c@{\hskip 0.05in}|}
\hline
Cars & Dresses & Pottery & Shoes & Natural & Small objects \\
\hline
0.60 & 0.84 & 0.76 & 0.80 & 0.81 & 0.74 \\
\hline
\end{tabular}
\caption{True positive ratio of adaptive ellipsoids for EPCS [Ps].}
\label{tab:daptation}
\end{table}

Our na\"{i}ve adaptation increases the true positive ratio by $1\%$ on three sets of real-world images (shoes and natural categories of Ebay dataset and small objects). This by no means is a finished adaptable model, rather a demonstration that our model can capture the variation in image content with the addition of simple extensions. This can be further explored by adapting chromatic ellipsoids to the presence or absence of certain colour categories in the image, following reports that link them to the phenomenon of colour constancy~\cite{vazquez2012color}. For instance when the green signal is abundant one could shrink the green ellipsoid or translate its centre. The adaptability of the ellipsoids in our model in turn could offer a framework in which colour constancy and colour categorisation are addressed simultaneously.

\section{Conclusion}

In this paper we presented a biologically-inspired colour categorisation model where each colour term is represented by an ellipsoid in colour-opponent space. To capture the fuzzy nature of colour names and account for the non-linear operations performed by visual cortex neurons, we computed the final degree of membership to a category using a sigmoid curve. Theoretically, we justified our geometrical framework by linking it to  physiological and psychophysical evidence. In practice, we showed that the parameters of our parsimonious model can be learnt from a simple optimisation procedure and conducted two kinds of experiments to verify its sanity. Results obtained on the Munsell chart are in excellent agreement with the psychophysical results of colour naming. We also perform better than other popular algorithms in real-world images. The advantage of the proposed model is more tangible by realising that, unlike all other state-of-the-art algorithms, it performs well on both types datasets. This shows that our model can both explain psychophysically-based colour naming results and perform an accurate categorisation of real-world images.

Biologically-inspired chromatic models have been successful in a wide range of colour computational tasks, e.g. colour induction~\cite{otazu2008multiresolution}, colour constancy~\cite{parraga2016colour,BMVC2017_331}, saliency~\cite{murray2011saliency} and boundary detection~\cite{BMVC2016_12,arash2017ijcv}. This is not surprising since colour is a sensation that originates from within our brains, which in turn is the product of millions of years of evolution, adaptation and ``learning'' from the visual environment. From this point of view, we believe our approach to colour categorisation can compete with other deterministic and learning-based approaches. In this line, we demonstrated that our model can be easily extended to incorporate more colour terms from few examples (as human infants do) and adapt itself to the content of image. Implicitly demonstrating the potential of biologically-inspired colour categorisation modelling for different applications such as image segmentation and image retrieval. Naturally, our model (as any other colour naming model) is likely to improve its accuracy in different environments when complemented with good colour constancy and colour induction algorithms and fundamentally with larger and better ground truths.

There are at this point two main lines of investigation for the future. The first one consists on improving the assignment of colour names by considering more sophisticated rules than a simple max pooling. The second one points to making the model dynamically responsive to context (either by rearranging the ellipsoids according to image content, or alternatively, supplementing the model with a centre-surround adaptation mechanism) to account for the well known colour phenomena of induction and constancy.

\section{Acknowledgement}

This work was funded by the Spanish Secretary of Research and Innovation (TIN2013-41751-P and TIN2013-49982-EXP) and has been partially presented at the European Conference on Visual Perception (ECVP)~\cite{akbarinia2015ecvp}.

{\small
\bibliographystyle{ieee}
\bibliography{ColourTermsArxivSupplementary}
}

\end{document}